\documentclass[conference]{IEEEtran}
\IEEEoverridecommandlockouts
\usepackage{hyperref}
\usepackage{cite}
\usepackage{amsmath,amssymb,amsfonts}
\usepackage{algorithmic}
\usepackage{graphicx}
\usepackage{textcomp}
\usepackage{xcolor}
\usepackage{booktabs}
\def\BibTeX{{\rm B\kern-.05em{\sc i\kern-.025em b}\kern-.08em
    T\kern-.1667em\lower.7ex\hbox{E}\kern-.125emX}}
\begin{document}

\title{SeqBench: Benchmarking Sequential Narrative Generation in Text-to-Video Models\\}

\author{
\IEEEauthorblockN{1\textsuperscript{st} Zhengxu Tang}
\IEEEauthorblockA{\textit{Department of Electrical} \\
\textit{and Computer Engineering}\\
\textit{University of Michigan}\\
Ann Arbor, United States \\
tangzx@umich.edu}
\and
\IEEEauthorblockN{1\textsuperscript{st} Zizheng Wang}
\IEEEauthorblockA{\textit{Department of Mechanical} \\
\textit{and Industrial Engineering}\\
\textit{Northeastern University}\\
Boston, United States \\
wang.zizheng@northeastern.edu}
\and
\IEEEauthorblockN{3\textsuperscript{rd} Luning Wang}
\IEEEauthorblockA{\textit{Department of Electrical}\\
\textit{and Computer Engineering}\\
\textit{University of Michigan}\\
Ann Arbor, United States \\
lnwang@umich.edu}
\and
\IEEEauthorblockN{4\textsuperscript{th} Zitao Shuai}
\IEEEauthorblockA{\textit{Department of Electrical} \\
\textit{and Computer Engineering}\\
\textit{University of Michigan}\\
Ann Arbor, United States \\
ztshuai@umich.edu}
\and
\IEEEauthorblockN{5\textsuperscript{th} Chenhao Zhang}
\IEEEauthorblockA{\textit{Paul G. Allen School of} \\
\textit{Computer Science and Engineering}\\
\textit{University of Washington}\\
Seattle, United States \\
cnhzhang@cs.washington.edu}
\and
\IEEEauthorblockN{6\textsuperscript{th} Siyu Qian}
\IEEEauthorblockA{\textit{School of Engineering} \\
\textit{ and Applied Sciences} \\
\textit{Harvard University}\\
Cambridge, United States \\
siyulilyqian@gmail.com}
\and
\IEEEauthorblockN{7\textsuperscript{th} Yirui Wu}
\IEEEauthorblockA{\textit{School of Electronic} \\
\textit{and Information Engineering}\\
\textit{Beijing Jiaotong University}\\
Beijing, China \\
21231083@bjtu.edu.cn}
\and
\IEEEauthorblockN{8\textsuperscript{th} Bohao Wang}
\IEEEauthorblockA{
\textit{College of Information Science}\\
\textit{and Electronic Engineering} \\
\textit{Zhejiang University}\\
Hangzhou, China \\
bohaowang@zju.edu.cn}
\and
\IEEEauthorblockN{9\textsuperscript{th} Haosong Rao}
\IEEEauthorblockA{
\textit{Georgen Institute for Data Science}\\
\textit{University of Rochester}\\
Rochester, United States \\
hrao@u.rochester.edu}
\and
\IEEEauthorblockN{10\textsuperscript{th} Zhenyu Yang}
\IEEEauthorblockA{
\textit{School of Earth Sciences}\\
\textit{Zhejiang University}\\
Hangzhou, China\\
3200101209@zju.edu.cn}
\and
\IEEEauthorblockN{11\textsuperscript{th} Chenwei Wu}
\IEEEauthorblockA{
\textit{Georgen Institute for Data Science}\\
\textit{University of Rochester}\\
Rochester, United States \\
cwu59@u.rochester.edu}
}
\maketitle

\begin{abstract}
Text-to-video (T2V) generation models have made significant progress in creating visually appealing videos. However, they struggle with generating coherent sequential narratives that require logical progression through multiple events. Existing T2V benchmarks primarily focus on visual quality metrics but fail to evaluate narrative coherence over extended sequences. To bridge this gap, we present SeqBench, a comprehensive benchmark for evaluating sequential narrative coherence in T2V generation. SeqBench includes a carefully designed dataset of 320 prompts spanning various narrative complexities, with 2,560 human-annotated videos generated from 8 state-of-the-art T2V models. Additionally, we design a Dynamic Temporal Graphs (DTG)-based automatic evaluation metric, which can efficiently capture long-range dependencies and temporal ordering while maintaining computational efficiency. Our DTG-based metric demonstrates a strong correlation with human annotations. Through systematic evaluation using SeqBench, we reveal critical limitations in current T2V models: failure to maintain consistent object states across multi-action sequences, physically implausible results in multi-object scenarios, and difficulties in preserving realistic timing and ordering relationships between sequential actions. SeqBench provides the first systematic framework for evaluating narrative coherence in T2V generation and offers concrete insights for improving sequential reasoning capabilities in future models. Please refer to \url{https://videobench.github.io/SeqBench.github.io/} for more details.
\end{abstract}


\begin{figure*}[t]
    \centering
    \includegraphics[width=\textwidth]{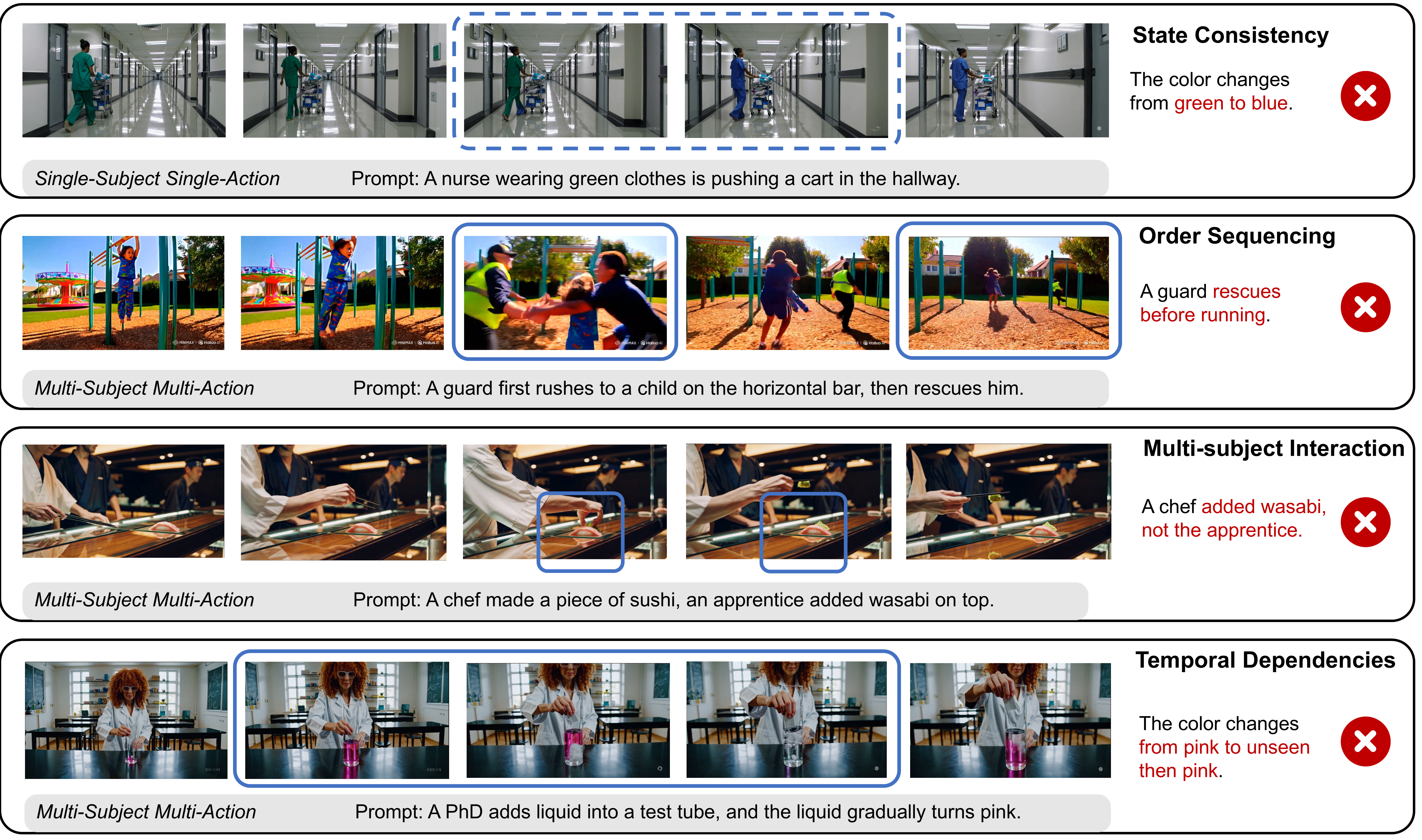}
    \caption{Examples of sequential narrative generation failures in current T2V models. Each row demonstrates a specific failure mode: State Consistency (nurse's clothing color changes), Order Sequencing (guard rescues before running), Multi-subject Interaction (wrong subject performs action), and Temporal Dependencies (incorrect liquid color transitions).}
    \label{fig:teaser}
\end{figure*}

\section{Introduction}
Text-to-video (T2V) generation models~\cite{videoworldsimulators2024} have demonstrated remarkable progress in synthesizing high-fidelity videos from textual descriptions~\cite{sun2024soraseesurveytexttovideo,liao2024evaluationtexttovideogenerationmodels}. While current approaches excel at producing visually compelling singular scenes, such as "a dog playing in the snow" or "a sunset over a city"~\cite{pika2024,kling2024}, how they perform on extended narratives that demand coherent progression through a sequence of actions and events remains underexplored. This limitation becomes increasingly critical as the T2V community advances toward real-world applications such as digital content creation~\cite{feng2024tc, meng2024towards}, where narrative coherence is as essential as visual appeal. 

Existing T2V benchmarks~\cite{qi2025t2veval, huang2024vbench++} primarily focus on visual aspects such as object fidelity, frame transitions, physical consistency and text-video alignment~\cite{zhang2024benchmarking,miao2024t2vsafetybench}. However, they overlook a fundamental aspect of human storytelling: how sequences of events unfold coherently over time. As shown in Fig.~\ref{fig:teaser}, even when given simple prompts, state-of-the-art models fail to preserve object states, follow logical event orders, reflect action consequences and handle multiple interacting entities. This inspires our first research question: \textbf{How good are current T2V models at generating coherent sequences of narratives?}

On the other hand, challenges arise for efficient and effective evaluation of long narratives. Recently, MLLM-based evaluation methods~\cite{wang2024your} have emerged as promising alternatives to laborious human evaluation. However, existing MLLM evaluation approaches face several critical limitations. First, most MLLMs struggle with processing entire videos due to computational constraints and model capability limits~\cite{llmcapabilitysurvey, llmperceptuallimit,qi2023mllmlimitation,efficientmllm}, making them unable to capture long-range dependencies and sequential ordering across extended narratives. Second, many evaluation methods~\cite{llmcapabilitysurvey,piper2024narritive,ye2025sequential} completely disregard sequential narrative structure and focus solely on frame-level assessment, missing the temporal coherence that defines good storytelling. Third, some approaches~\cite{tan2024keyframe, su2025distillingtransitional} rely heavily on key-frame detection and may miss crucial transitional details between actions. Lastly, more recent methods like StoryEval~\cite{wang2024your} only consider event completion rates while completely ignoring event ordering, consequence accuracy, and logical consistency between actions. This brings our second research question: \textbf{How do we design effective automatic evaluation methods for both sequential narrative coherence and visual details?}

To address this evaluation gap, we introduce SeqBench, a comprehensive framework for evaluating both narrative coherence and visual fidelity in T2V tasks involving sequences of events. We constructed a diverse benchmark dataset comprising 320 carefully designed prompts spanning various narrative complexities, with 2,560 human-labeled videos generated from 8 state-of-the-art T2V models. Additionally, to enable automatic and systematic evaluation, we devise a novel evaluation metric leveraging MLLMs to understand narrative sequences through Dynamic Temporal Graphs (DTG). DTG-based evaluation is more comprehensive in capturing fine-grained details and long-term dependencies while being computationally efficient compared to full video processing approaches. We validate our metric against human evaluation, showing strong correlation ($\rho$=0.857) and demonstrating its reliability for automatic assessment. Through this framework, we are able to systematically analyze previously underexplored challenges in sequential narrative generation, including action sequencing and transition quality, multi-object interaction consistency, and sequential dependency preservation across extended temporal spans. Our benchmark can be publicly accessed at \url{https://huggingface.co/datasets/AcmmmVideobench/Acmmm2025_video_benchmark}.

Through comprehensive evaluation of leading T2V models using SeqBench, we identify critical limitations in current approaches: Most models fail to maintain consistent object states across multi-action sequences; multi-object scenarios frequently produce physically implausible or sequentially inconsistent results; and while individual actions may appear fluid, models struggle with maintaining realistic timing and ordering relationships between sequential or concurrent actions.

In short, we contribute: 
\begin{itemize}
    \item A \textbf{new benchmark} specifically designed to evaluate sequential narrative coherence in T2V generation. 
    \item A \textbf{new metric} that efficiently captures long-range dependencies and temporal ordering.
    \item A \textbf{new dataset} of 320 prompts and 2,560 video-text pairs with detailed annotations and human evaluations.
    \item \textbf{New insights} revealing current T2V models' limitations in sequential reasoning, providing concrete directions for model improvements.
\end{itemize}
\section{Dataset}
\subsection{Overview}
To comprehensively evaluate whether text-to-video (T2V) models can generate long narratives across diverse scenarios, we design our benchmark around three dimensions: \textbf{content categories} that capture different themes, \textbf{difficulty levels} that systematically probe models' capabilities in handling scenes of varying complexity, and \textbf{temporal orders} that specify how actions take place relative to each other in time.
\subsubsection{Content Categories}
We establish four fundamental content categories that encompass the breadth of scenarios commonly encountered in video generation:
\textbf{Animal} category which focuses on animal behaviors and interactions, ranging from simple locomotion to complex predatory or social behaviors; \textbf{Human} category which encompasses human activities across various contexts, from basic daily routines to complex social interactions; \textbf{Object} category that centers on inanimate objects and their transformations, movements, or interactions with other entities; \textbf{Imaginary} category that includes fantastical, supernatural, or highly stylized content that extends beyond realistic constraints.
\subsubsection{Difficulty Levels}
We define four increasing difficulty levels structured around the number of subjects and the number of actions involved in the narrative sequence.
\textbf{Single Subject-Single Action (SSSA)} is the baseline difficulty involving one subject performing a single, self-contained action.\textbf{ Single Subject-Multi Action (SSMA)} involves one subject performing multiple actions across an extended sequence. This tests models' ability to maintain temporal ordering and state transitions across consecutive actions. \textbf{Multi Subject-Single Action (MSSA)} involves multiple subjects collectively performing or participating in the same action, requiring consistent tracking of multiple entities while maintaining their individual characteristics and coordinated behavior. \textbf{Multi Subject-Multi Action (MSMA)}  requires multiple subjects to perform different actions across complex sequences, often involving coordination and dependencies between subjects. This tests complex coordination and inter-subject temporal relationships.
\subsubsection{Temporal Orders}
We define three temporal orders that specify how actions unfold relative to each other within each difficulty level.
\textbf{Strictly Sequential (SS)} requires actions to follow a predetermined, logical sequence where each action must complete before the next begins. This tests models' ability to maintain strict temporal ordering and clear state transitions. \textbf{Flexible Order (FO)} allows actions to occur in varying orders while maintaining logical coherence, probing models' understanding of arbitrary sequencing. \textbf{Simultaneous (SI)} challenges models with concurrent actions, testing their ability to coordinate parallel processes and maintain consistency across simultaneous state changes.

In summary, our dataset contains 4 main categories (Animal, Human, Object, Imaginary) and 4*8 = 32 subcategories (SSSA, MSSA SSMA-SS, SSMA-FO, SSMA-SI, MSMA-SS, MSMA-FO, MSMA-SI). 
\subsection{Prompt Suite Construction}
We design each prompt in our benchmark to feature 1–4 logical actions or events that can unfold within a short time window (since most publicly available T2V models can only generate videos<=10 seconds). To maintain consistency and clarity, we utilize standardized prompt templates that defines the scene setting, character description, and action chain. 
\subsubsection{3-Step Prompt Generation Process}
 First, our \textbf{Retrieval and Captioning} approach involves selecting existing short video clips that exemplify our target action patterns across animal, human, object, and imaginary scenarios. We manually transform these visual sequences into structured text prompts while preserving their temporal-sequential relationships and ensuring they align with our difficulty level specifications. Second, our \textbf{Human Brainstorming} process adds both realistic and imaginative scenarios while maintaining adherence to basic sequential logic. This approach allows us to explore creative boundaries, particularly within our imaginary category, while ensuring that even fantastical scenarios maintain internal consistency and logical progression. Third, our \textbf{LLM-Driven Generation} process guides large language models to produce concise multi-step action descriptions that satisfy our temporal-sequential requirements and difficulty specifications. These steps yield 200 raw prompts for each subcategory, which is a total of 6400 prompts.
\subsubsection{Prompt Filtering} 
\begin{figure}
    \centering
    \includegraphics[width=1\linewidth]{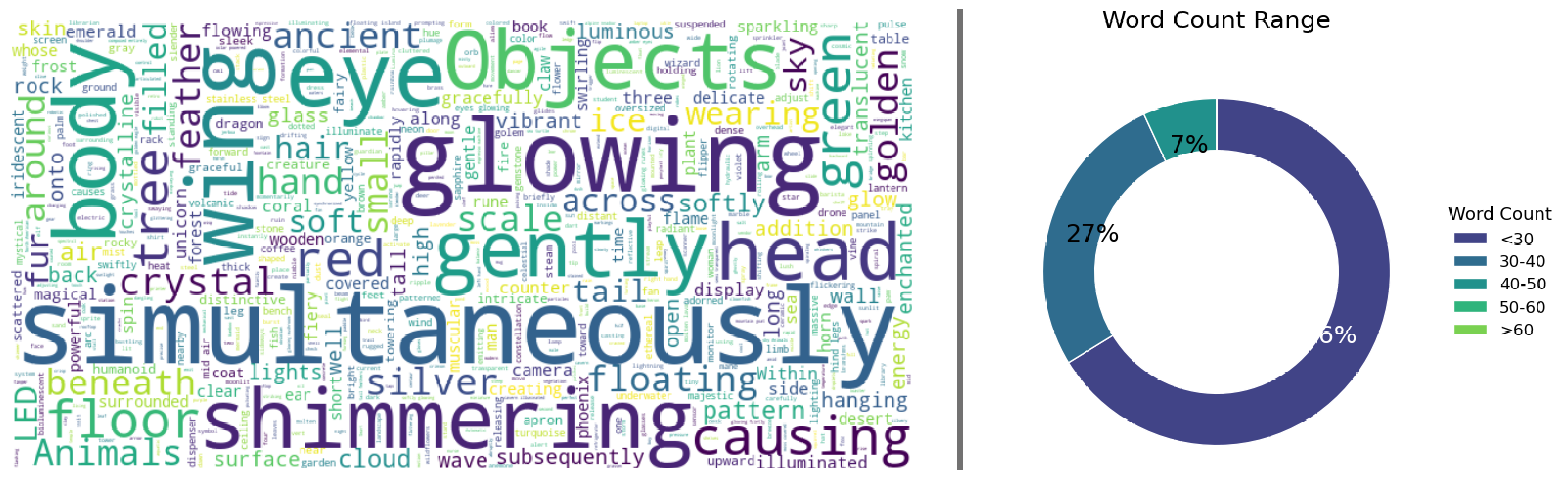}
    \caption{Dataset statistics and prompt characteristics. Left: Word cloud visualization showing the most frequently occurring terms across all 320 prompts in SeqBench, with word size indicating frequency. Right: Distribution of prompt lengths by word count ranges.}
    \label{fig:stats}
\end{figure}

We then apply rigorous filtering criteria to ensure the quality and feasibility of our prompt suite. Specifically, the chosen prompts should satisfy: 1) \textbf{Appropriate Length}, where all actions within each prompt must be completable within the specified 1-5 second duration, and the prompts should not exceed 77 tokens, which is the limit of the CLIP text encoders adopted by most T2V models; 2) \textbf{Audio-Free Descriptions}, where no references to sounds, music, or audio elements should be included in the prompts, to focus exclusively on visual narrative elements; 3) \textbf{Scenario Adherence}, where the prompts strictly conform to their assigned content category and maintain consistency with the specified difficulty level and temporal order; and 4) \textbf{No Jargons and Over-detailed Descriptions}, where the prompts should avoid too complex or hyper-specific descriptions that exceed current T2V model capabilities, focusing on achievable yet challenging sequential narratives. This step completes our prompt suite construction, resulting in 320 high-quality prompts evenly distributed across each subcategory. Dataset visualization can be seen in Fig. \ref{fig:stats}.







\section{Evaluation Metrics}

\subsection{Overview}

To address the limitations of existing evaluation approaches for sequential narrative coherence, we propose a novel automatic evaluation framework that assesses videos across two complementary dimensions: \textbf{Visual Details Evaluation} and \textbf{Narrative Coherence Evaluation}. Our method systematically evaluates both visual fidelity and temporal consistency, providing comprehensive assessments of text-to-video generation quality.

\subsection{Visual Details Evaluation}

The visual details evaluation pipeline assesses the quality and accuracy of visual content generation using frame-level analysis.

\textbf{Frame-level Scene Graph Extraction}: We first extract scene graphs from representative frames to capture the visual elements present in the video. The scene graph represents objects, their attributes, and spatial relationships in a structured format.

\textbf{Question Generation On Video Content}: We then generate evaluation questions focused on visual accuracy, including object presence, attribute correctness, spatial relationship accuracy, and scene composition quality. These questions assess whether the generated video correctly represents the visual elements described in the text prompt.

\textbf{Frame-based Question Answering}: We finally use visual question answering techniques to evaluate the extracted scene graphs against the generated questions, providing scores for visual fidelity and content accuracy.

\subsection{Narrative Coherence Evaluation}

\begin{figure*}[t]
    \centering
    \includegraphics[width=\textwidth]{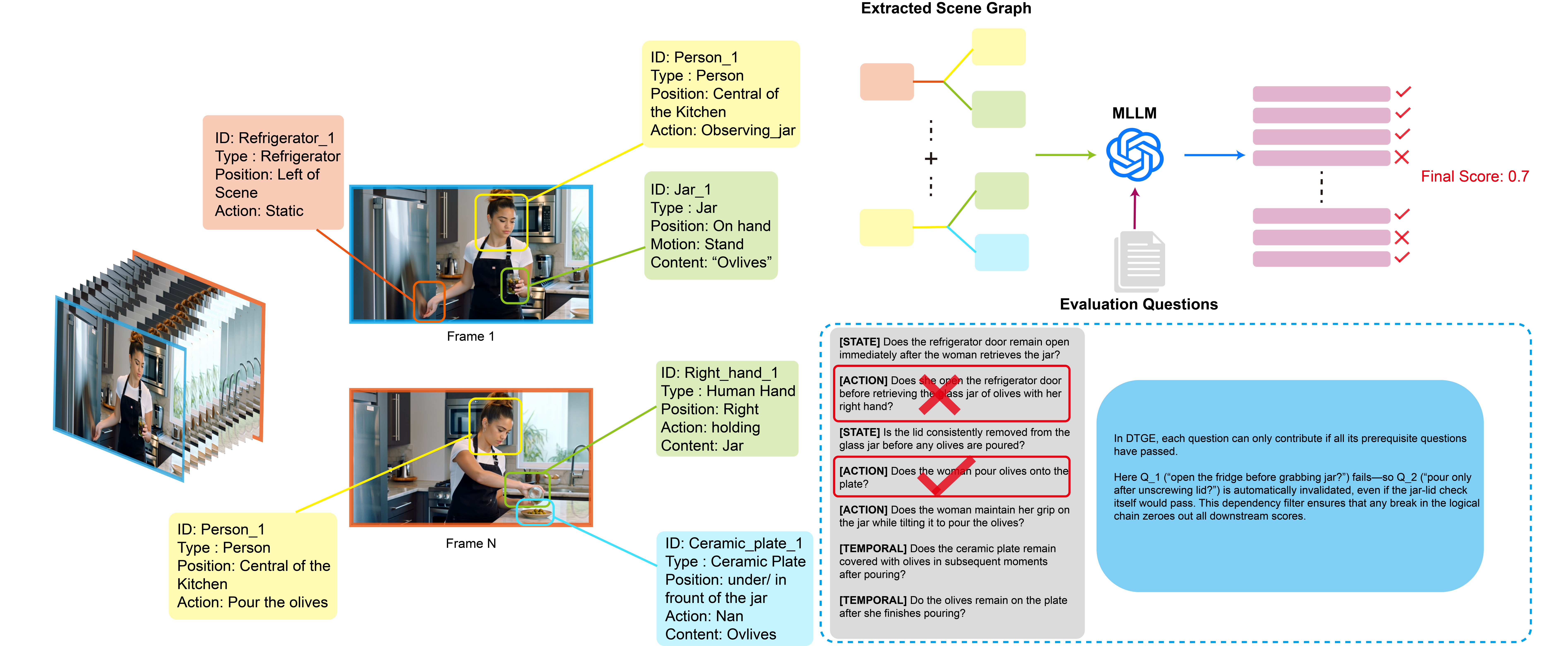}
    \caption{Overview of the Dynamic Temporal Graph (DTG) evaluation framework. The method extracts scene graphs from multiple video frames (left), processes them through the DTG pipeline to generate evaluation questions covering state consistency, action sequencing, and temporal dependencies (center), applies dependency filtering and scoring mechanisms (bottom right), and computes final coherence scores. }
    \label{fig:metric}
\end{figure*}

The narrative coherence evaluation pipeline assesses temporal consistency and sequential logic using Dynamic Temporal Graphs (DTG), as illustrated in Fig. \ref{fig:metric}.

\subsubsection{Temporal Decomposition}

Given a video prompt describing a sequence of events, we first decompose it into structured temporal components. We parse the input prompt into temporal tuples of the form \texttt{(id, type, content)}, where each tuple represents either a state transition or an action event. We then establish temporal dependencies between tuples to capture logical relationships and sequential constraints, creating a directed acyclic graph representing the logical flow of the narrative.

\subsubsection{Dynamic Graph Extraction}

Traditional scene graph extraction uses static templates that may miss narrative-specific details. We introduce \textbf{adaptive graph extraction} that dynamically adjusts extraction prompts based on the specific evaluation questions for each video.

\textbf{Dynamic Prompt Generation}: The core innovation lies in generating customized graph extraction prompts for each video based on its specific evaluation questions. Our system analyzes the generated questions to identify which visual features, object states, and temporal changes are most critical for evaluation. It then modifies the base graph extraction prompt to emphasize tracking of these specific elements. For instance, if evaluation questions focus on object state changes (e.g., ``Does the cup's fullness level properly transition?''), the adapted prompt will specifically instruct the vision model to pay closer attention to container objects and their fill states across frames.

\textbf{Question-aware Feature Emphasis}: The adaptation process works by parsing the evaluation questions to extract key entities, actions, and state transitions mentioned in each question. These extracted elements are then integrated into the graph extraction instructions, ensuring that the vision model prioritizes detecting and tracking the exact features needed for accurate evaluation. This targeted approach significantly improves the relevance and accuracy of extracted scene graphs compared to generic extraction methods.

\textbf{Multi-frame Graph Extraction}: We extract scene graphs from multiple frames (typically 15 frames) distributed across the video timeline using the adapted prompts. Each graph captures objects and entities with attributes including position, action states, visual properties, spatial and semantic relationships between entities, and associated confidence scores. The adaptive prompting ensures that crucial narrative elements are consistently tracked across all frames, enabling robust temporal analysis.

\subsection{Scoring and Aggregation}

\textbf{Visual Details Scoring}: Frame-level evaluation produces scores for visual accuracy and content fidelity, indicating how well the generated video represents the visual elements described in the text prompt.

\textbf{Narrative Coherence Scoring}: Each temporal question receives a binary score (1.0 for ``yes'', 0.0 for ``no''). The question will only be counted as correct when all questions related to prior action/objects are correct, a process we refer to as 'dependency filtering'. Please refer to Fig. \ref{fig:metric} and \url{https://videobench.github.io/SeqBench.github.io/} for more details.


The combination of visual details and narrative coherence scores provides a comprehensive assessment of text-to-video generation quality, capturing both visual fidelity and temporal consistency.

\subsection{Human Evaluation}
We conduct an extensive human evaluation on all video-prompt pairs in our benchmark to validate that our DTG-based evaluation score aligns with human judgment. We recruited 11 experienced participants to perform human evaluation on the 2 dimensions, visual Quality and Narrative Coherence.
Both evaluation tasks employ a 3-point Likert scale (0, 1, 2), where 0 indicates poor quality/coherence, 1 represents acceptable performance, and 2 denotes excellent quality/coherence. To ensure consistency and reliability, each video is independently evaluated by 5 different participants, and we report the average score across all evaluators.
\begin{table*}[t]
\centering
\caption{DTG Metric - Visual Details Evaluation Results}
\resizebox{\textwidth}{!}{%
\begin{tabular}{l*{8}{c}*{4}{c}c}
\toprule
\multicolumn{1}{c}{Model}
  & \multicolumn{8}{c}{Difficulty–time order combinations}
  & \multicolumn{4}{c}{Content categories}
  & \multicolumn{1}{c}{Avg.\ Score} \\
\cmidrule(lr){2-9} \cmidrule(lr){10-13}
 & SSMA\_SS & SSMA\_FO & SSMA\_SI & MSMA\_SS & MSMA\_FO & MSMA\_SI & SSSA & MSSA 
 & Animal & Human & Imaginary & Object 
 &  \\
\midrule
Runway Gen3     & 0.485 & 0.552 & 0.563 & 0.383 & 0.446 & 0.442 & 0.532 & 0.470 
         & 0.530 & 0.465 & 0.436 & 0.505 
         & 0.484 \\
Sora     & 0.737 & 0.764 & 0.750 & 0.549 & 0.591 & 0.608 & 0.699 & 0.693 
         & 0.749 & 0.646 & 0.610 & 0.691 
         & 0.674 \\
Luma Ray2    & 0.684 & 0.681 & 0.724 & 0.518 & 0.608 & 0.543 & 0.721 & 0.676 
         & 0.683 & 0.624 & 0.610 & 0.660 
         & 0.644 \\
Veo 2.0     & 0.482 & 0.585 & 0.595 & 0.523 & 0.517 & 0.476 & 0.627 & 0.578 
         & 0.677 & 0.527 & 0.464 & 0.523 
         & 0.548 \\
Cogvideo 1.5 & 0.806 & 0.817 & 0.814 & \textbf{0.700} & 0.718 & 0.711 & \textbf{0.780} & 0.789 
         & 0.811 & \textbf{0.806} & 0.701 & 0.749 
         & 0.767 \\
Pika 2.2     & 0.757 & 0.762 & 0.721 & 0.565 & 0.676 & 0.595 & 0.710 & 0.733 
         & 0.745 & 0.723 & 0.600 & 0.705 
         & 0.690 \\
Hailuo T2V-01  & 0.812 & 0.826 & 0.758 & 0.654 & 0.659 & 0.650 & 0.771 & 0.780 
         & 0.790 & 0.726 & 0.715 & 0.724 
         & 0.739 \\
Kling 2.0    & \textbf{0.819} & \textbf{0.839} & \textbf{0.823} & 0.693 & \textbf{0.734} & \textbf{0.713} & 0.774 & \textbf{0.804} 
         & \textbf{0.821} & 0.778 & \textbf{0.739} & \textbf{0.763} 
         & \textbf{0.775} \\
\bottomrule
\end{tabular}%
}
\label{tab:visual}
\end{table*}

\begin{table*}[t]
\centering
\caption{DTG Metric - Narrative Coherence Evaluation Results }
\resizebox{\textwidth}{!}{%
\begin{tabular}{l*{8}{c}*{4}{c}c}
\toprule
\multicolumn{1}{c}{Model}
  & \multicolumn{8}{c}{Difficulty–time order combinations}
  & \multicolumn{4}{c}{Content categories}
  & \multicolumn{1}{c}{Avg.\ Score} \\
\cmidrule(lr){2-9} \cmidrule(lr){10-13}
 & SSMA\_SS & SSMA\_FO & SSMA\_SI & MSMA\_SS & MSMA\_FO & MSMA\_SI & SSSA & MSSA 
 & Animal & Human & Imaginary & Object 
 &  \\
\midrule

Runway Gen3     & 0.149 & 0.137 & 0.163 & 0.115 & 0.141 & 0.163 & 0.198 & 0.170 
         & 0.184 & 0.150 & 0.159 & 0.123 & 0.154 \\
Sora     & 0.198 & 0.161 & 0.154 & 0.123 & 0.152 & 0.171 & 0.287 & 0.180 
         & 0.226 & 0.169 & 0.175 & 0.143 & 0.178 \\
Luma Ray2     & 0.122 & 0.174 & 0.149 & 0.171 & 0.199 & 0.146 & 0.216 & 0.262 
         & 0.237 & 0.161 & 0.199 & 0.146 & 0.180 \\
Veo 2.0    & 0.156 & 0.180 & 0.158 & 0.184 & 0.225 & 0.190 & 0.256 & 0.262 
         & 0.286 & 0.178 & 0.173 & 0.167 & 0.201 \\
Cogvideo 1.5 & 0.222 & 0.197 &\textbf{ 0.208} & 0.189 & 0.224 & 0.191 & 0.272 & 0.235 
         & 0.289 & 0.221 & 0.197 & 0.161 & 0.217 \\
Pika 2.2    & 0.219 & 0.202 & 0.194 & 0.171 & 0.238 & \textbf{0.229} & 0.293 & 0.277 
         & 0.278 & 0.214 & 0.219 &\textbf{ 0.200} & 0.228 \\
Hailuo T2V-01   & 0.212 & \textbf{0.230} & 0.192 & 0.179 & 0.207 & 0.199 & \textbf{0.345} & 0.288 
         & 0.298 & 0.238 & 0.215 & 0.175 & 0.231 \\
Kling 2.0    & \textbf{0.266} & 0.209 & 0.207 & \textbf{0.192} &\textbf{ 0.258} & 0.228 & 0.291 & \textbf{0.366 }
         & \textbf{0.324} &\textbf{ 0.259} &\textbf{ 0.240} & 0.185 & \textbf{0.252 }\\
\bottomrule
\end{tabular}
}
\label{tab:coherence}
\end{table*}

\section{Results}
In this section, we evaluate 8 popular video generative models on our SeqBench benchmark. We consider Runway Gen3~\cite{runway2024gen3alpha}, Sora~\cite{videoworldsimulators2024}, Luma Ray2~\cite{luma2024ray2}, Google Veo 2.0~\cite{google2024veo2}, CogVideo 1.5~\cite{yang2024cogvideox}, Pika 2.2~\cite{pika2024}, Hailuo T2V-01~\cite{minimax2024hailuo}, and Kling 2.0~\cite{kling2024}. More details about these models are provided in the project page.
\subsection{How Good Are Current T2V Models at Generating Extended Sequences of Narratives?}
Our evaluation reveals a striking discrepancy in current T2V model capabilities. As shown in Table \ref{tab:visual}, at the frame level, all models demonstrate excellent visual quality, generating high-fidelity individual frames with realistic and prompt-matching details, lighting, and object appearances. Leading performers like Kling 2.0 (average score: 0.775), Cogvideo 1.5 (average score: 0.767) consistently produce visually appealing content across all scenarios.
However, when it comes to video-level narrative coherence, performance drops dramatically across all models. As shown in Table \ref{tab:coherence}, even the best-performing model (Kling 2.0) achieves only a 0.252 average coherence score, indicating substantial room for improvement in sequential narrative generation. The performance gap between visual quality and narrative coherence highlights a fundamental limitation: while models excel at generating beautiful individual frames, they struggle significantly with maintaining logical progression across multi-action sequences.
Kling 2.0 emerges as the clear leader, demonstrating superior performance across most difficulty-temporal order combinations, particularly excelling in single-subject multi-action scenarios (SSMA\_SS: 0.266) and multi-subject multi-action scenarios (MSSA\_FO: 0.366). Notably, Kling shows resilience even in challenging scenarios where other models fail completely, often producing the most coherent results despite technical imperfections.
\subsection{Are Our Evaluation Metrics Sufficiently Aligned with Human Judgment?}
\begin{table}[ht]
\centering
\caption{Spearman $\rho$ values between DTG score and human evaluation.}
\label{tab:merged_spearman}
\begin{tabular}{lcc}
\hline
Category     & $\rho$ Visual Quality & $\rho$ Narrative Coherence \\ 
\hline
SSMA\_SS     & 0.627                  & 0.695                  \\ 
SSMA\_FO     & 0.762                  & 0.619                  \\ 
SSMA\_SI     & 0.557                  & 0.776                  \\ 
MSMA\_SS     & 0.739                  & 0.778                  \\ 
MSMA\_FO     & 0.786                  & 0.970                  \\ 
MSMA\_SI     & 0.881                  & 0.667                  \\ 
SSSA         & 0.619                  & 0.761                  \\ 
MSSA         & 0.814                  & 0.838                  \\ 
\hline
\hline
\end{tabular}
\end{table}
To validate the reliability of our DTG-based evaluation framework, we measure the correlation between our automatic metrics and human evaluations using Spearman's rank correlation coefficient $\rho$. As shown in Table\ref{tab:merged_spearman}, our evaluation metrics demonstrate strong alignment with human judgment across both visual quality and narrative coherence dimensions.
\subsection{What Are the Common Pitfalls of Current T2V Generation Models?}
We present insights based on both automatic evaluations using our Dynamic Temporal Graph Evaluation (DTGE) metric and qualitative observations collected through human review.
\subsubsection{Complexity-Driven Performance Degradation}
Performance drops sharply as prompts demand complex actions or involve multiple subjects—especially in the "Multi-Subject, Multi-Action" (MSMA) categories. Multi-subject scenarios fundamentally complicate the generation process, with models struggling most when strict sequential ordering is required (MSMA\_SS scores consistently lowest across all models). Interestingly, while multi-subject scenarios remain challenging, simultaneous actions (MSMA\_SI) prove slightly more manageable than strict sequential ordering, suggesting models find concurrent actions easier to coordinate than maintaining precise temporal sequences. 
\subsubsection{Subject Inconsistency and Omission}
A pervasive issue across models is the tendency to miss or inconsistently render subjects, particularly in multi-subject scenarios. Models frequently generate fewer subjects than specified in the prompt, with occasional instances of generating more subjects than intended. This problem is especially pronounced in lower-quality models like Sora and Gen3, which show significant subject count errors. For simultaneous actions, models tend to employ "deceptive" strategies when faced with multi-subject scenarios—instead of showing all subjects in the same frame, they often resort to sequential cuts, displaying object A for the first few seconds, then switching to object B, and finally object C, creating a "scene-switching" effect rather than true simultaneous action.
\subsubsection{Temporal Logic and Action Coordination Failures}
Models exhibit fundamental limitations in both sequential and simultaneous action scenarios. In logical multi-action sequences (SSMA\_SS/MSMA\_SS), we observe frequent temporal reversals—showing effects before causes, such as juice spilling before a blender tips over. Complex multi-step actions requiring precise coordination often result in incomplete execution, with models generating only one or two actions from longer sequences, demonstrating poor prompt utilization and action dependency understanding. In simultaneous scenarios (MSMA\_SI/SSMA\_SI), models frequently employ "action dropping," omitting actions or entire subjects rather than tracking multiple concurrent processes, defaulting to simpler single-focus narratives.
\subsubsection{Model-Specific Behaviors}
Gen3 excels at generating natural scenes with believable motion but frequently ignores prompt specifications, inventing its own narrative instead of following the given instructions. Sora and Pika share similar human rendering styles but consistently produce anatomical distortions, including extra limbs and warped environmental elements.
Veo 2.0 demonstrates strong performance but suffers from technical artifacts including repetitive playback, frame jumping, and sudden character/object disappearances that disrupt narrative flow.

\section{Conclusion}
This work introduces SeqBench, the first comprehensive benchmark for evaluating narrative coherence in T2V generation. Through systematic evaluation, we reveal a fundamental limitation: while models excel at visual quality, they struggle significantly with narrative coherence.
Our novel DTG evaluation framework demonstrates strong correlation with human judgment while efficiently capturing long-range temporal dependencies. We identify critical failure patterns including object state inconsistency, multi-subject tracking failures, temporal logic reversals, and incomplete action execution—providing concrete directions for future model improvements.

\bibliographystyle{IEEEtran}
\bibliography{IEEEfull}

\end{document}